\documentclass{article}

\usepackage{microtype}
\usepackage{graphicx}
\usepackage{subfigure}
\usepackage{booktabs} 

\usepackage{hyperref}

\usepackage[accepted]{icml2024}

\usepackage{amsmath}
\usepackage{amssymb}
\usepackage{mathtools}
\usepackage{amsthm}

\usepackage[capitalize,noabbrev]{cleveref}

\theoremstyle{plain}
\newtheorem{theorem}{Theorem}[section]

\theoremstyle{definition}

\theoremstyle{remark}
\newtheorem{remark}[theorem]{Remark}
\newcommand{\blockcomment}[1]{}

\usepackage[textsize=tiny]{todonotes}

\icmltitlerunning{Gradient Boosted Filters}

\graphicspath{ {./figures/} }

\begin{document}

\twocolumn[
\icmltitle{Gradient Boosted Filters For Signal Processing}




\begin{icmlauthorlist}
\icmlauthor{Jose A. Lopez}{comp}
\icmlauthor{Georg Stemmer}{comp}
\icmlauthor{Héctor A. Cordourier}{comp}
\end{icmlauthorlist}

\icmlaffiliation{comp}{Intel Labs, Intel Corp.}

\icmlcorrespondingauthor{Jose A. Lopez}{jose.a.lopez@intel.com}

\icmlkeywords{gradient boosting, signal processing, volterra}

\vskip 0.3in
]



\printAffiliationsAndNotice{}  

\begin{abstract}
Gradient boosted decision trees have achieved remarkable success in several domains, particularly those that work with static tabular data. However, the application of gradient boosted models to signal processing is underexplored. In this work, we introduce gradient boosted filters for dynamic data, by employing Hammerstein systems in place of decision trees. We discuss the relationship of our approach to the Volterra series, providing the theoretical underpinning for its application. We demonstrate the effective generalizability of our approach with examples.
\end{abstract}

\section{Introduction}
\label{sec:introduction}
Gradient boosting models (GBMs) employ a sequence of weak learners with each learner correcting the prediction errors of its predecessor. This correction is attained by setting the target for a learner to the prediction error of the predecessor during training. The final prediction is obtained by summing all the predictions. The theory does not specify a type of loss or the type of weak learner and this underscores the flexibility of the approach. While the theory behind GBMs does not prescribe a specific type of learner, decision trees have become the de facto choice in practice due to their performance and interpretability and many implementations that use decision trees are available  \cite{interpret_2019,catboost_2018,lightgbm_2017,xgboost_2016}.

Besides being versatile, GBMs are renowned for their efficiency and high performance on tabular data, often surpassing deep learning methods that require significantly more computational resources. However, the inherently static nature of decision trees hampers their ability to model dynamic data, where the temporal relationships between data points are crucial. This limitation motivates the exploration of alternative weak learners that can capture such dynamics without cumbersome data augmentation to make temporal information explicit. Given the performance, lightweight nature, ease of training, and relative explainability of GBMs, we believe it is worth exploring their potential beyond static data applications.

In this study, we expand the gradient boosting framework to effectively accomodate dynamic data.  We propose the novel integration of linear filters as part of the weak learner. However, linear filters alone cannot function as the weak learner. As stated by the projection theorem, the prediction error of the optimized first filter will be orthogonal to the range space of subsequent filters. Thus, subsequent stages are unable to improve the overall prediction. We need something more. We proceed this way, from first principles, to develop gradient boosted filters (GBFs) that employ Hammerstein systems in place of decision trees. Our journey leads us to an unexpected intersection with an established and rich domain of signal processing, specifically to the Volterra and Wiener series. Given the equivalence of the Wiener and Volterra series and the latter's more intuitive notation, our discussion and subsequent methodology will focus on the Volterra framework.

The paper is organized as follows: we begin with a concise review of the Volterra series to set the stage for our proposed GBFs. We then detail our methodology and present experimental results that highlight the potential of our approach to broaden the applicability of GBMs. We conclude with a discussion of our findings and potential avenues for future research.

\section{Volterra Series}
\label{sec:volterra_series}
The Volterra series is often characterized in two ways: as a Taylor series with memory \cite{drongelen_2018,schetzen_chapter_2010,mathews_1993,boyd_dissertation_1985}, or as a generalization of the linear time-invariant (LTI) operator to non-linear operators \cite{moodi_2010,boyd_1984}. Following the notation in \cite{schetzen_chapter_2010}, the series expansion is written as:

\begin{equation}
y(t) = h_0 + \sum_{n=1}^\infty H_n[x(t)]
\label{eq:volterra_series}
\end{equation}
where $y(t)$ is the output of a continuous non-linear system, $x(t)$ is the input, and 
\small
\begin{align}
H_n[x(t)] = \nonumber\\ 
\int_{-\infty}^{\infty} \cdots \int_{-\infty}^{\infty} h_n(\tau_1,\dots,\tau_n) 
x(t-\tau_1)\dots x(t-\tau_n) d\tau_1 \dots d\tau_n \nonumber \\
\label{eq:volterra_operator}
\end{align}
\normalsize
We call $H_n$ the $n$-th order Volterra operator and $h_n$ the Volterra kernel of $H_n$. Each $h_n$ can be considered the system response to multiple impulses at different times and is not generally unique unless it is symmetric \cite{drongelen_2018,boyd_1984}. The $0$-th Volterra kernel $h_0$ is a constant that can be disregarded in our development. 

From Equations \ref{eq:volterra_series} and \ref{eq:volterra_operator}, one can see that both descriptions commonly used to describe the Volterra series are apt. However, we lean towards the first description as it encapsulates the pros and cons of the Volterra series which are inherited from the Taylor series. To summarize, the Volterra series can approximate non-linear systems arbitrarily closely, albeit with restrictions on discontinuities \cite{mathews_1993}. For a more comprehensive explanation, see \cite{drongelen_2018,schetzen_chapter_2010, boyd_dissertation_1985,palm_1977}.

In practical applications, the discrete time description is preferred, and we utilize the discrete expansion with operators defined as:

\small
\begin{align}
H_n[x(t_n)] = \nonumber \\
\sum_{\tau_1=0}^{T_1} \cdots \sum_{\tau_n=0}^{T_n} h_n(\tau_1,\dots,\tau_n) x(t_n-\tau_1)\dots x(t_n-\tau_n)
\label{eq:discrete_volterra_operator}
\end{align}
\normalsize
Here, it is noteworthy that the sum is now causal and truncated. Limiting the sum to $T_n$ implies finite system memory. Fortunately, the class of causal, non-linear, systems with finite memory is extensive. Moreover, each $h_n$ has dimension $n$. This introduces the primary obstacle in obtaining a Volterra series description: identification of the kernels. This issue has been the subject of numerous studies and it is generally difficult to identify the kernels \cite{wray_1994}. Consequently, most works that identify Volterra series representations for non-linear systems limit the order to a few operators \cite{drongelen_2018}.

\subsection{Related Work}
\label{ssec:related_work}

Several authors have implemented Volterra representations using neural networks. In \cite{hakim_1991}, the authors delineate the Volterra kernels of a single-layer neural network and a recurrent network, discussing the systems that can be represented using such networks. The networks are fed temporal information through delayed samples at the input, making the input size dependent on the system being modeled, which is generally the case. In Section \ref{sec:simple_dynamical_system}, we utilize their example of a dynamical system to demonstrate the ability of GBFs to generalize to out-of-distribution data.

In \cite{ahmed_1991}, the authors introduce the Nadine architecture, capable of modeling any Volterra series. Nadine is constructed using layers of adaptive linear filters, which utilize the outputs of one layer as weights for the next, rather than the input. Consequently, the output is generated by a linear filter with state-dependent weights. Specifically, the output of Nadine can be written as:

\begin{equation}
y(n) = w_0 + \sum_{i=0}^{N-1} \alpha_{i}(\mathbf{x}) x(n-i)
\label{eq:nadine_y}
\end{equation}
where $\mathbf{x}$ is a vector of the input with past samples, or the state. The $\alpha$'s are determined by another linear filter with state dependent weights:
\begin{equation}
\alpha_{i}(\mathbf{x}) = w_{i} + \sum_{j=0}^{N-1} \beta_{i,j}(\mathbf{x}) x(n-j)
\label{eq:nadine_alpha}
\end{equation}
and the $\beta$'s are the outputs of the input filter with weights that are state independent:
\begin{equation}
\beta_{i,j}(\mathbf{x}) = w_{i,j} + \sum_{k=0}^{N-1} w_{i,j,k} x(n-k)
\label{eq:nadine_beta}
\end{equation}

The state-dependent weights change due to the alterations in the outputs of the previous layer and the local adaptation of the weights not included in the sum. The entire network is trained using separate (local) applications of the least-mean-squared (LMS) update rule. The authors propose that this local learning reduces approximation error and improves training speed. However, in \cite{osowski_1996,osowski_1994}, the authors employ the adjoint flow graph to derive update formulas for the weights, allowing the approximation error to update the entire network. Our GBF approach merges non-iterative local learning with backpropagation, and each stage has the same input but is trained with a different target.

In \cite{wray_1994}, the authors demonstrated that a simple feed-forward network with delayed inputs is equivalent to a discrete Volterra series and derived the expressions for its kernels. The study provided a sufficient condition: if such a network can approximate a non-linear system, then the associated kernels can be extracted using the formulas provided. In Section \ref{sec:laptop_acoustics}, we compare with our implementation of the feed-forward network described by the authors. 

More recently, \cite{moodi_2010} proposed that the kernels could be replaced by Laguerre functions and wavelets in a Volterra series-like expression to model a hydraulic actuator using fewer parameters. Our approach similarly departs from the focus on imposing a kernel architecture, prioritizing usability instead. 

In a recent study by \cite{banerjee_2021}, the authors presented their generalization of Volterra convolutional neural networks to Riemannian homogeneous spaces, with applications to non-Euclidean data found in fields such as computer vision and medical imaging.

The study of \cite{gallman_1975}, which is perhaps the most similar to ours but does not employ neural networks, uses a parallel sum of Hammerstein systems to identify non-linear systems and shows convergence for a class of inputs, including colored Gaussian processes. However, the algorithm does not estimate the static non-linearities as they are assumed to be known. Additionally, each stage improves the residual error between the target and the sum of the outputs of all the other stages, meaning the order of the stage is not important. In contrast, in our work, each stage builds upon the previous stages, aiding practitioners in determining the number of stages required for their problem, which is crucial for preventing over-fitting.

\section{Gradient Boosted Filters}
\label{sec:gbf}
In our quest for an optimal weak learner for GBFs, we searched for a learner capable of modeling non-linear dynamic information in a straightforward way. We found Hammerstein systems, which are composed of a static non-linearity followed by a linear filter, to be well-suited for this purpose. While the key difference between Hammerstein and Wiener systems is the order of the components, we opted for Hammerstein systems as they allow for closed-form estimation of the linear part, thereby reducing the number of parameters requiring estimation via backpropagation. 

\begin{figure}[h]
\centering
\includegraphics[width=0.7\columnwidth]{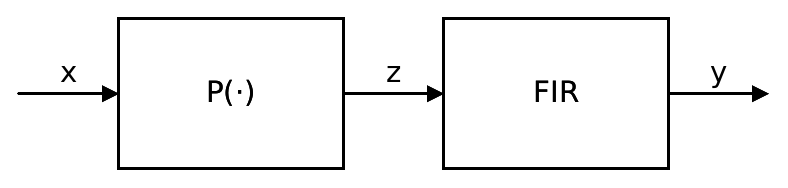}
\caption{A Hammerstein system with a polynomial non-linearity and FIR component.}
\label{fig:weak_learner}
\end{figure}

In GBFs, all stages have the same form, shown in Figure \ref{fig:weak_learner}, but can be independently adjusted for efficiency. Specifically, the $i$-th stage of the GBF is given by Equation \ref{eq:gbf_stage}.

\begin{equation}
y_i(n) = \sum_{j=0}^{m_i} \sum_{k=0}^{p_i} b_{ij} a_{ik} x(n-j)^k
\label{eq:gbf_stage}
\end{equation}
where each $b_{ij}$ is a coefficient of a causal finite impulse response (FIR) filter of length $m_i$, and each $a_{ik}$ is a coefficient of a polynomial of order $p_i$. Each stage is trained to predict the difference between the output of the previous stage and its target. Therefore, the target for the $i$-th stage is given by $t_i=t_{i-1}-y_{i-1}$. The final prediction is obtained by summing the individual outputs $y_i$.

From Equation \ref{eq:gbf_stage}, it is evident that the cross-products between delayed inputs from the Volterra series are absent, replaced by a polynomial expansion of an individual delayed term. If we used a Wiener formulation, the polynomial would be computed using the state of the FIR filter, and the cross terms would be present. This would demonstrate equivalence to the truncated Volterra series as was argued in \cite{wray_1994}. However, we would not be able to optimally estimate the FIR filter in closed form via the Wiener-Hopf equations, thereby reducing the number of parameters estimated using backpropagation. Therefore, we believe the polynomial expansion is expressive enough to improve the approximation error at each stage and proceed with Hammerstein systems for implementation. While we focus on the regression problem, with the mean squared error (MSE) loss, the extension to classification is straightforward and we leave it for future work.

\begin{remark}[Expressiveness.]
The Weierstrass approximation theorem enables us to assert that any static non-linearity can be trivially modeled by GBFs using a single stage. This is not the case for Volterra series \cite{roheda_2019}. To see this, set the FIR of a single-stage GBF to a Kronecker delta and select a polynomial to have $p_i$ as high as needed. However, the use of FIR filters comes with a trade-off: the systems that can be modeled by GBFs cannot exhibit an infinite response to an impulse. Alternatively, the terms must be negligible beyond the number of coefficients available for the application.
\end{remark}

Algorithm \ref{alg:algorithm_separate} gives the steps used to train a GBF model in a stage-wise manner. However, when there are sufficient computational resources available, it is feasible to train the polynomials of all the stages simultaneously. This is done by back-propagating the cumulative prediction errors as shown in Algorithm \ref{alg:algorithm_combined}.

\begin{algorithm}
   \caption{Gradient Boosted Filters (separate training)}
   \label{alg:algorithm_separate}
\begin{algorithmic}
   \STATE {\bfseries Input:} data $x$, target $t$, FIR lengths $m_i$, polynomial orders $p_i$, number of learners $N$.
   \STATE Initialize target $t_1 = t$.
   \FOR{$i=1$ {\bfseries to} $N$}   
   \REPEAT
   \STATE{1. Transform input.
   \begin{equation}   
   z_i = \sum_{k=0}^{p_i} a_{ik} x^k
   \end{equation}
   }
   \STATE{
   2. Compute Wiener-Hopf filter coefficients
   \begin{equation}
   b_i = E[t_i z_i^T] E[z_iz_i^T]^{-1}
   \end{equation}
   where $E[\cdot]$ denotes expectation.
   }
   \STATE{
   3. Compute prediction $y_i$ and residual $r_i=t_i - y_i$.
   }
   \STATE{
   4. Backpropagate error $\|r_i\|_2^2$ to update $a_{ik}$.
   }
   \UNTIL{$\|r_i\|_2^2$ has converged.}
   \STATE{
   5. Set target for next learner $t_{i+1} = r_i$.   
   }
   \ENDFOR

\end{algorithmic}
\end{algorithm}

\begin{remark}[Wiener-Hopf solution.]
While we have included the time-domain Wiener-Hopf solution in Algorithms \ref{alg:algorithm_separate} and \ref{alg:algorithm_combined} to emphasize that the optimal least-squares estimate relies solely on the auto-correlation of the filter input and cross-correlation of the filter input and desired output, it is more practical to utilize the fast Fourier transform (FFT) to solve for the coefficients in the frequency domain.

In many scenarios, such as in the audio domain, the number of FIR coefficients significantly exceeds that of the polynomial terms. Consequently, using the Wiener-Hopf solution substantially reduces the number of coefficients that need to be learned via back-propagation. This implies a significant reduction in the risk of over-fitting, enhancing the generalization capability of the model.
\end{remark}

\begin{algorithm}
   \caption{Gradient Boosted Filters (combined training)}
   \label{alg:algorithm_combined}
\begin{algorithmic}
   \STATE {\bfseries Input:} data $x$, target $t$, FIR lengths $m_i$, polynomial orders $p_i$, number of learners $N$.
   \STATE Initialize target $t_1 = t$.
   \REPEAT
   \FOR{$i=1$ {\bfseries to} $N$}   
   
   \STATE{1. Transform input.
   \begin{equation}   
   z_i = \sum_{k=0}^{p_i} a_{ik} x^k
   \end{equation}
   }
   \STATE{
   2. Compute Wiener-Hopf filter coefficients.
   \begin{equation}
   b_i = E[t_i z_i^T] E[z_iz_i^T]^{-1}
   \end{equation}
   }
   \STATE{
   3. Compute prediction $y_i$ and residual $r_i=t_i - y_i$.
   }     
   \STATE{
   4. Set target for next learner $t_{i+1} = r_i$.   
   }
   \ENDFOR
   \STATE{
   5. Backpropagate error $\sum_i \|r_i\|_2^2$ to update $a_{ik}$.
   } 
   \UNTIL{$\sum_i \|r_i\|_2^2$ has converged.}

\end{algorithmic}
\end{algorithm}

In the following section, we will showcase the modeling capabilities of GBF models through examples. Based on our experiments, we found Algorithm \ref{alg:algorithm_separate} to be beneficial for hyperparameter exploration, while Algorithm \ref{alg:algorithm_combined} proved to be more effective for generating final predictions, often yielding better results. 

\section{Experiments}
\label{sec:experiments}
The subsequent experiments were run on a HP ZBook Fury 15 G7 laptop, equipped with an Intel i7-10850H CPU, 32 GB of memory, and an NVIDIA Quadro RTX 3000 GPU with 6.144 GB of memory. The software for this work was developed using the PyTorch framework.

\subsection{Example 1: A simple dynamical system}
\label{sec:simple_dynamical_system}
We employ an example from \cite{hakim_1991} to demonstrate the generalization capabilities of GBFs. In \cite{hakim_1991}, it was observed that the performance of the trained recurrent neural network deteriorated when the inputs deviated from the training distribution. The system is specified by the following equations:

\begin{align}
x(n+1) &= 0.5 x(n) + u(n) + 0.1 u(n)^2\\
t(n) &= x(n)
\label{eq:simple_dynamical_system}
\end{align}

For training, we sampled $200$ input points uniformly from the interval $[-1,1]$, a validation set was drawn uniformly from $[-2,2]$, and a test set from $[-4,4]$. The outputs were computed using Equation \ref{eq:simple_dynamical_system}.

\begin{figure}[h]
\centering
\includegraphics[width=\columnwidth]{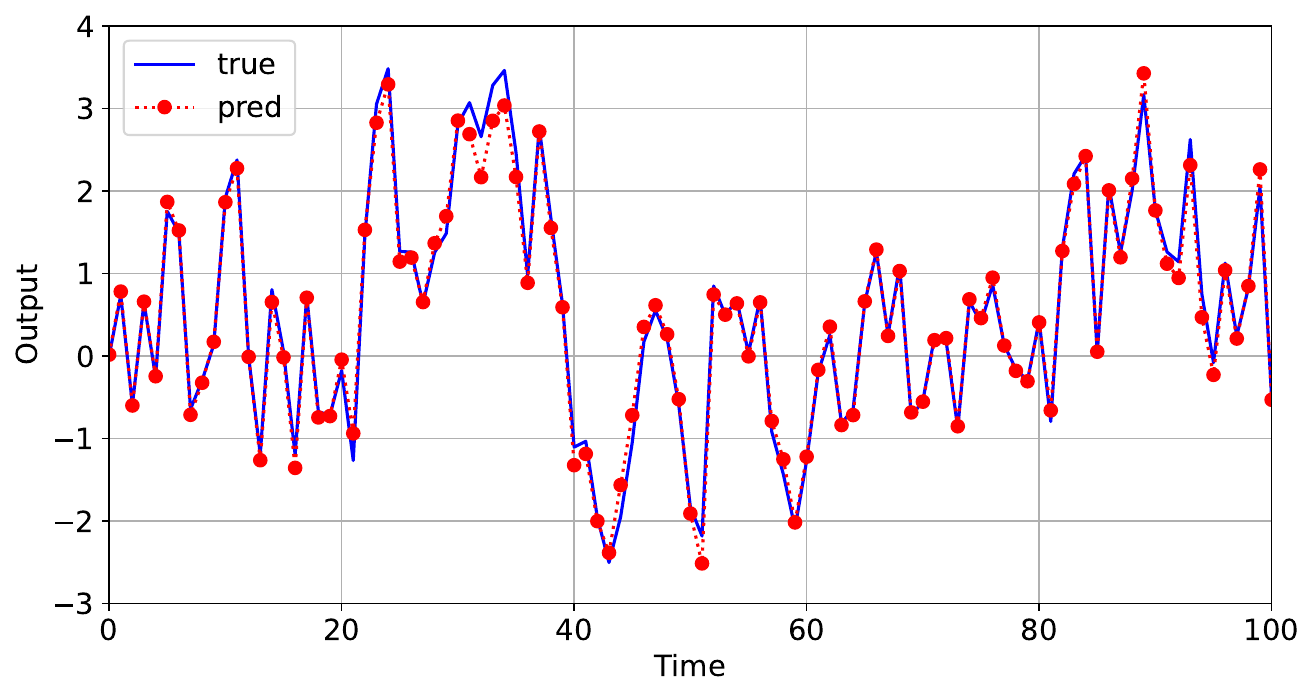}
\caption{Model output for input data with values between -2 and 2.}
\label{fig:ex1_val_fit}
\end{figure}

\begin{figure}[h]
\centering
\includegraphics[width=\columnwidth]{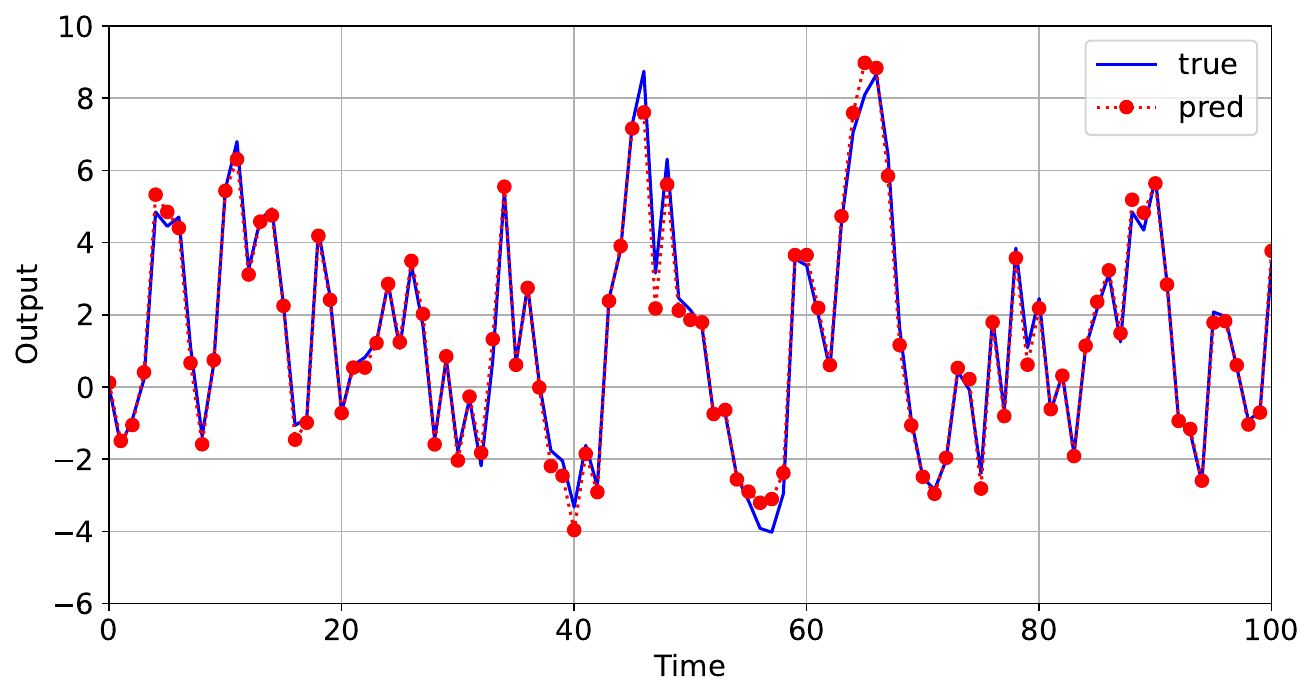}
\caption{Model output for input data with values between -4 and 4.}
\label{fig:ex1_test_fit}
\end{figure}

For the GBF, we utilized three stages with $(p_1,m_1)=(1, 24)$, $(p_2, m_2)=(2, 16)$, and $(p_3, m_3)=(2, 8)$. Consequently, the first filter is linear, and the next two incorporate quadratic non-linearities. Figures \ref{fig:ex1_val_fit} and \ref{fig:ex1_test_fit} demonstrate that the GBF is able to generalize to input values beyond the range found in the training data, outperforming the model from \cite{hakim_1991}.

\begin{figure}[h]
\centering
\includegraphics[width=\columnwidth]{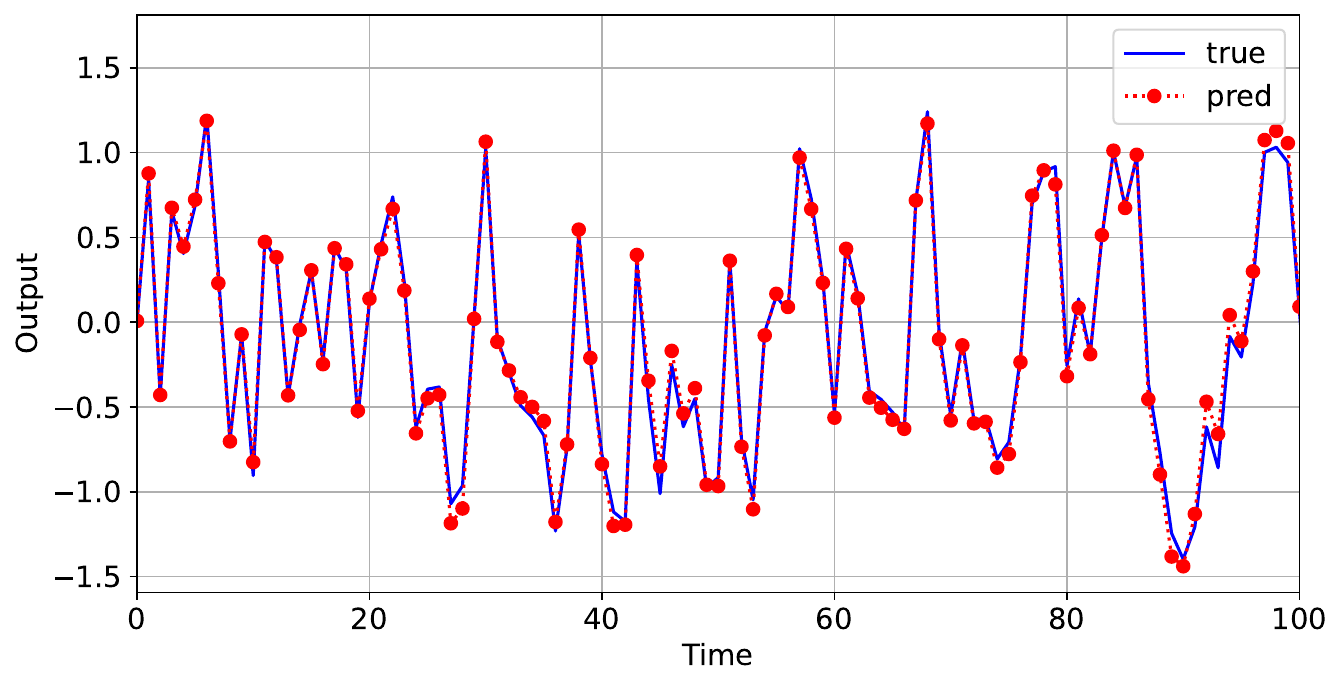}
\caption{Model output for input data with values between -1 and 1.}
\label{fig:ex1_train_fit}
\end{figure}

\subsection{Example 2: Laptop Acoustics}
\label{sec:laptop_acoustics}
In this example, we consider the problem of modeling laptop microphone and speaker acoustics within a conference room using a recording of chirps played through the speakers of the laptop and captured by its microphone. A laptop manufacturer might be interested in such a model to enhance noise or echo cancellation algorithms by leveraging the acoustic model, potentially by generating synthetic recordings to fine-tune existing algorithms for a new laptop.

A conventional approach for this task involves identifying a room impulse response (RIR) model from the chirp recordings and using it to create synthetic data. However, this linear model may fall short of capturing non-linearities effectively, thereby limiting the utility of the recordings.

\begin{figure}[h]
\centering
\includegraphics[width=\columnwidth]{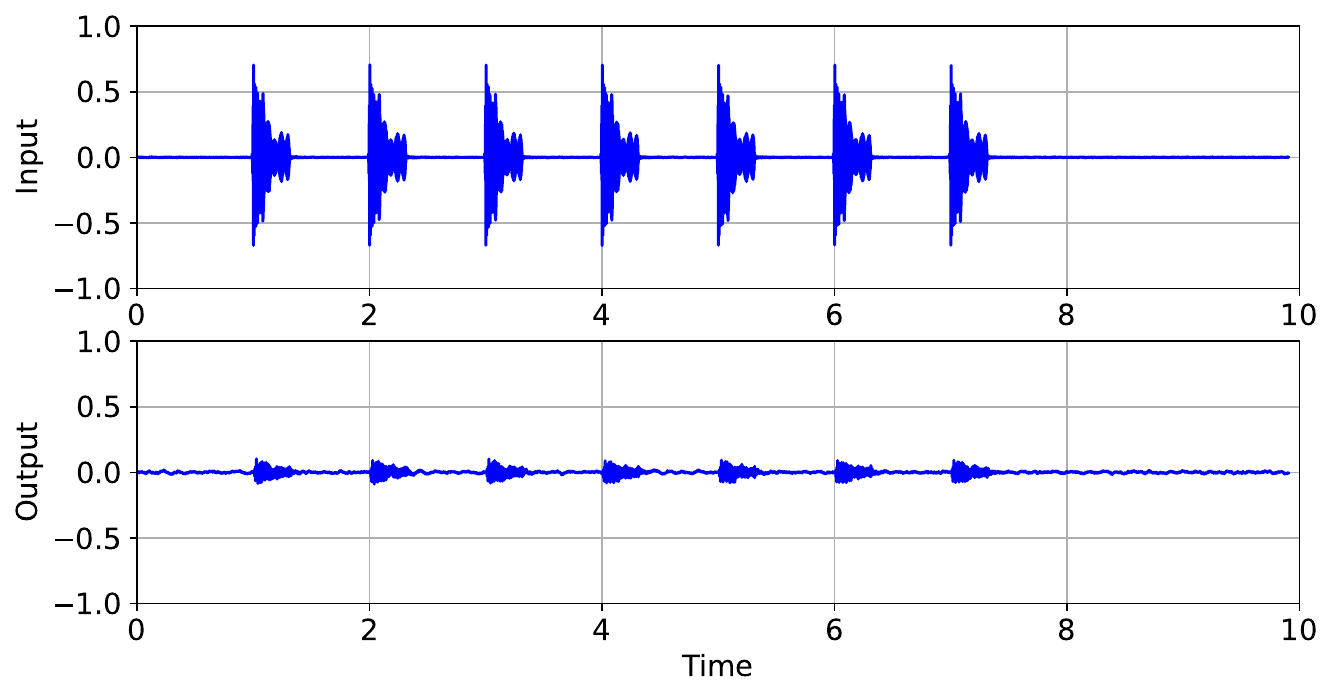}
\caption{Reference and recorded (16 KHz) audio signals.}
\label{fig:ex2_io_signals}
\end{figure}

Our aim is to identify a model from a single sequence of chirps using GBFs. This task presents a significant challenge, not only due to the modeling aspect but also because it essentially constitutes a one-shot learning problem. Figure \ref{fig:ex2_io_signals} displays the reference chirps played by the speakers and the signal recorded by the microphone. Notably, there are just seven chirps. We will utilize the first six for training and reserve the last for validation. For a blind test set, we utilize a 30-second clip from the song ``Monkeys Spinning Monkeys'' \cite{macleod_2019}. In absence of ground truth data, the output power from the test can serve as a distortion measure and compared against the optimal baseline FIR filter that does not introduce distortion.

\begin{figure}[h]
\centering
\includegraphics[width=\columnwidth]{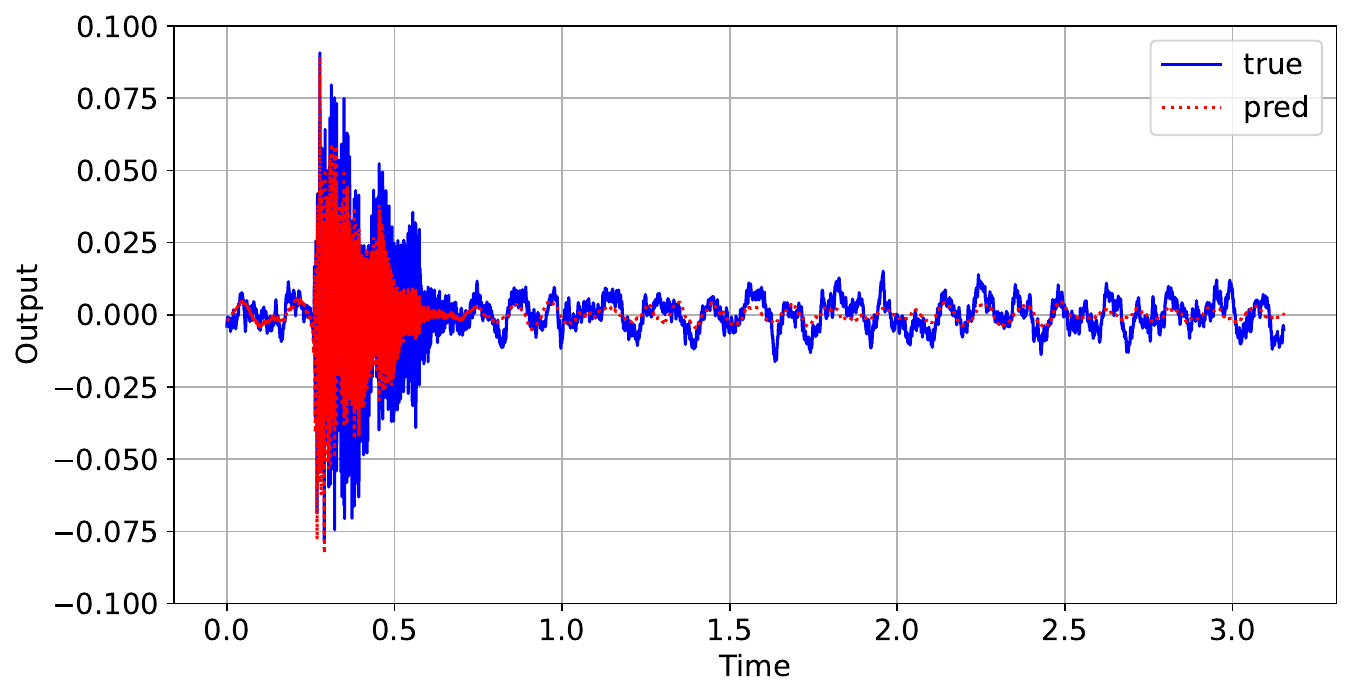}
\caption{The output generated by the baseline linear model with 6400 coefficients. The model attains MSEs of $5.44\times 10^{-5}$ and $4.77\times 10^{-5}$ on the training and validation data, respectively, serving as a performance reference for subsequent models.}
\label{fig:ex2_baseline_val}
\end{figure}

As a benchmark, we solve the Wiener-Hopf equations to obtain an optimal FIR filter with 6400 coefficients. The number of coefficients is approximately the duration of a single chirp. The output generated by this baseline model is shown in Figure \ref{fig:ex2_baseline_val}.

\begin{figure}[h]
\centering
\includegraphics[width=\columnwidth]{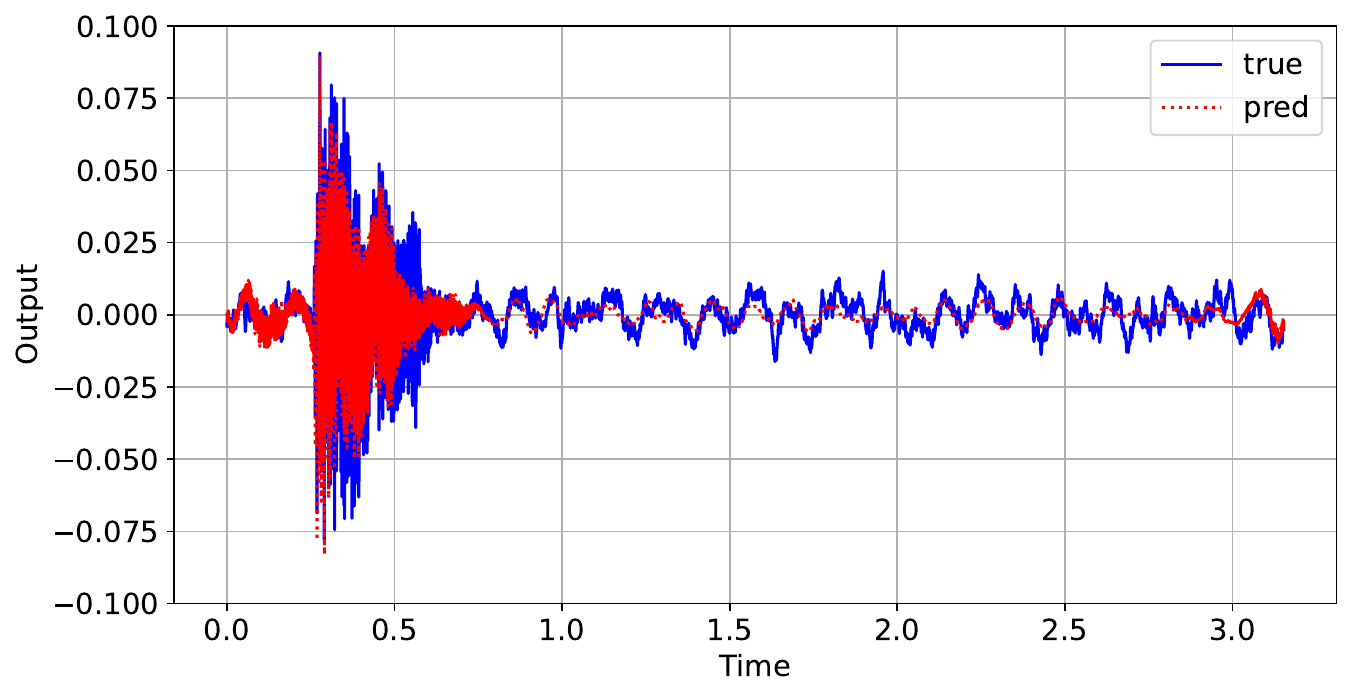}
\caption{The output generated by the GBF model, achieving MSEs of $2.86\times 10^{-5}$ and $3.16\times 10^{-5}$ on the training and validation data, resp. This represents a $47.44\%$ and $33.84\%$ improvement over the baseline model.}
\label{fig:ex2_gbf_val}
\end{figure}

For the GBF, we train an 11-stage model with $(p_1, m_1)=(1,6400)$ and $(p_i, m_i)=(10,6400)$ for $i\in[2,11]$. Therefore, the first stage mirrors the baseline model. This model has 110 polynomial parameters and 70,400 filter coefficients. Figure \ref{fig:ex2_gbf_val} displays the predicted validation chirp output of the GBF model. 

\begin{figure}[h]
\centering
\includegraphics[width=\columnwidth]{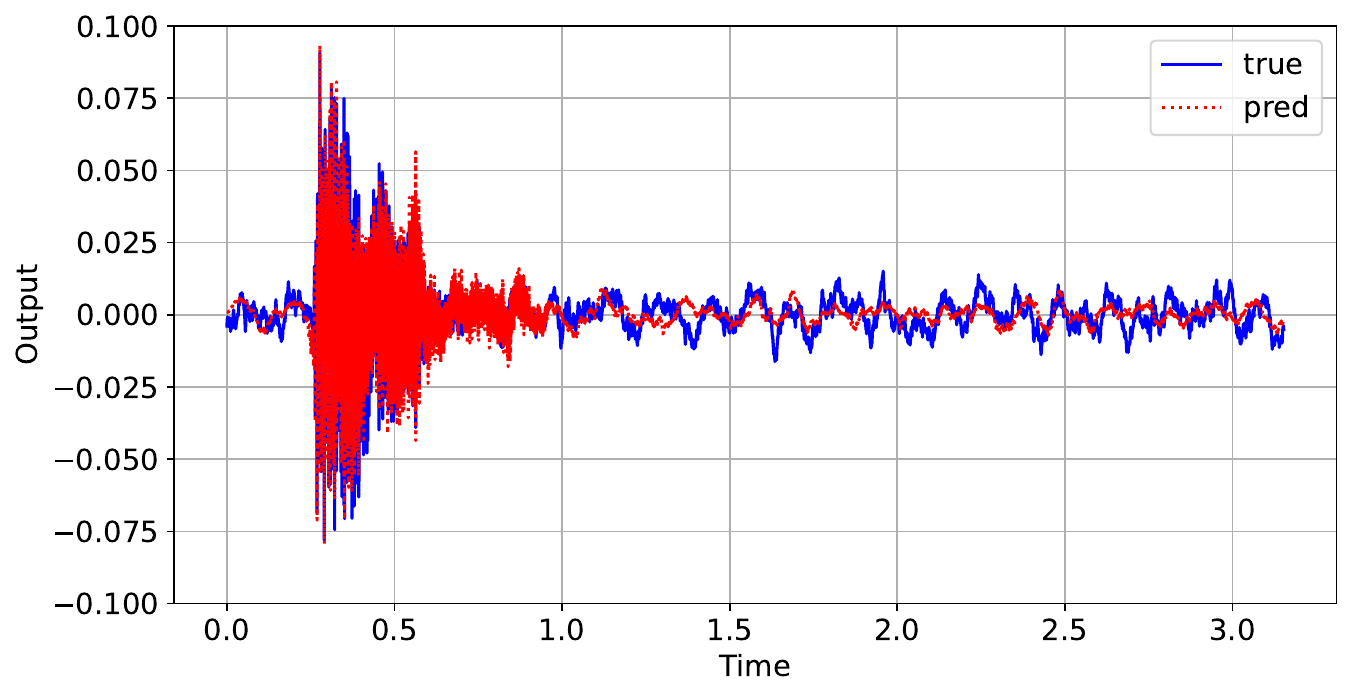}
\caption{The output generated by the Volterra model, attaining MSEs of $1.48\times 10^{-5}$ and $3.11\times 10^{-5}$ on the training and validation data, resp. This represents a $72.85\%$ and $34.81\%$ improvement over the baseline model.}
\label{fig:ex2_volterra_val}
\end{figure}

\begin{figure}[h]
\centering
\includegraphics[width=\columnwidth]{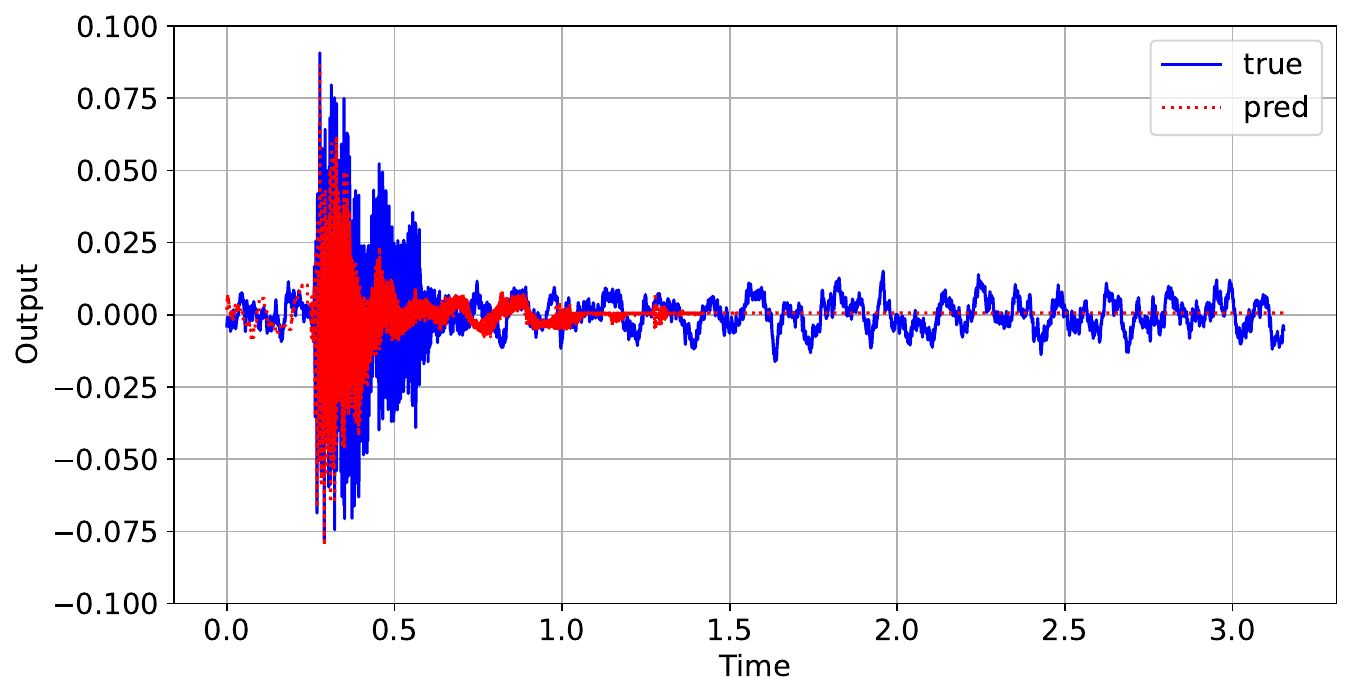}
\caption{The output of the TCN model, attaining MSEs of $6.38\times 10^{-5}$ and $5.25\times 10^{-5}$ on the training and validation data, resp. This signifies a $17.37\%$ and $10.15\%$ degradation from the baseline model.}
\label{fig:ex2_tcn_val}
\end{figure}

\begin{figure}[h]
\centering
\includegraphics[width=\columnwidth]{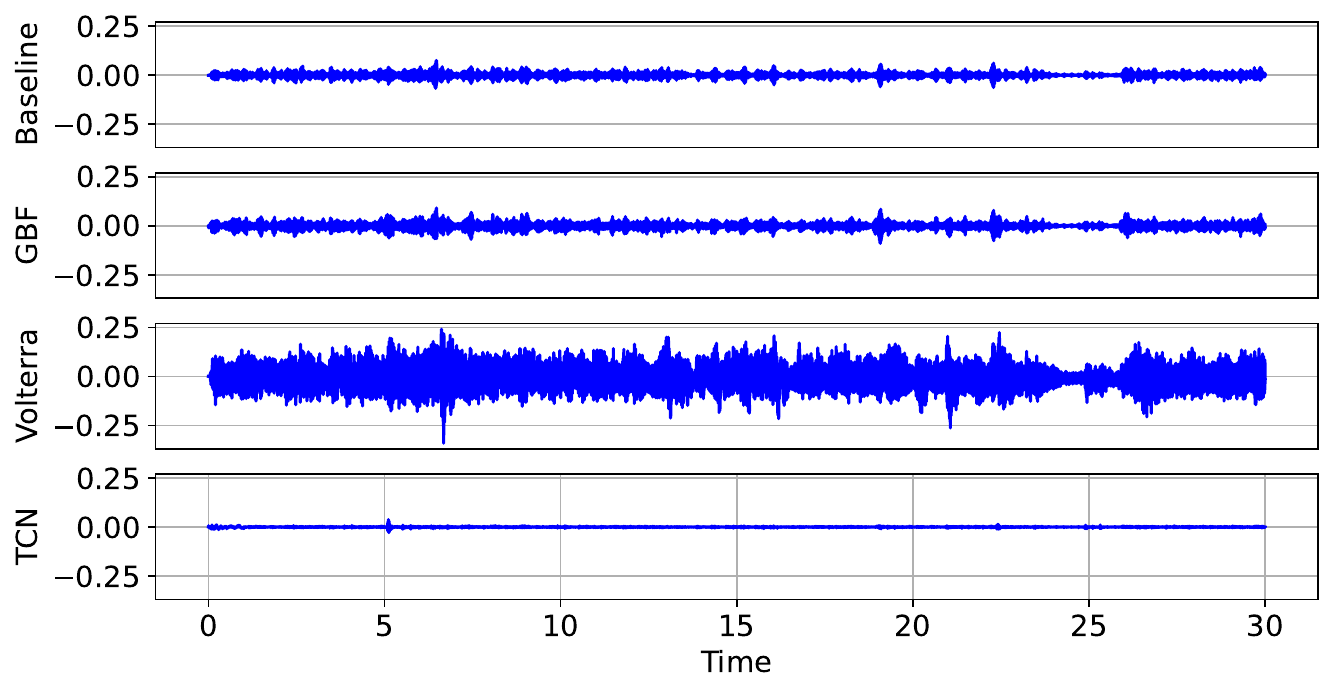}
\caption{Model outputs for the blind test: baseline and GBF models show similar output power.}
\label{fig:ex2_blind_test}
\end{figure}

For comparison, we train a Volterra network similar to the model presented in \cite{wray_1994}, maintaining the same temporal context as the GBF and baseline filter. Additionally, we also train a Temporal Convolutional Network (TCN) model, which is currently considered a state-of-the-art method for time series data \cite{khandelwal_2021,bai_2018}. The TCN model is configured with 13 levels, 32 hidden units, and a kernel size of 5, giving it a larger receptive field than the other models. The layers of the Volterra model are detailed in Table \ref{tab:volterra_architecture}, while the architecture and implementation of the TCN can be found in \cite{bai_2018}. The Volterra network has 76,825 parameters, while the TCN model has 129,921 parameters. All three models were trained using identical configurations: employing the AdamW optimizer with weight decay, a one-cycle scheduler, and for 18,000 epochs. Figure \ref{fig:ex2_volterra_val} displays the predicted validation chirp output of the Volterra model, while Figure \ref{fig:ex2_tcn_val} presents the predicted validation chirp output of the TCN model. It is clear the TCN performance is inferior to even the baseline. Figure \ref{fig:ex2_blind_test} demonstrates that, despite the Volterra network and the GBF model having similar size and validation performance, the GBF has superior generalization.

\begin{table}[H]
\caption{The architecture of the Volterra model.}
\label{tab:volterra_architecture}
\vskip 0.15in
\begin{center}
\begin{small}
\begin{sc}
\begin{tabular}{|c|}
\hline
Volterra model\\
\hline
causal Conv1d(1, 12, 6400) \\
\hline
LeakyReLU \\
\hline
Conv1d(12, 1, 1)\\
\hline
\end{tabular}
\end{sc}
\end{small}
\end{center}
\vskip -0.1in
\end{table}

\subsection{Conclusions}
\label{sec:conclusions}
Despite the astounding success of deep neural networks across various domains, there remain areas where smaller models are favored. This preference is particularly evident in domains with poor signal-to-noise ratios, such as stock trading and data mining, where smaller models tend to generalize better. In this paper, we have presented our extension of the GBM framework for dynamic data. We propose that GBFs can provide practitioners with an additional tool for efficiently modeling dynamic data. The generalizability of GBFs, we believe, can be attributed to the fact that FIR filters do not introduce distortion to the input signal. By employing Hammerstein systems that include FIR filters as the linear component, practicioners can regulate the degree of non-linearity, and consequently, the amount of distortion, thereby enabling greater control over the bias versus variance trade-off.

Looking ahead, our primary focus will be on generalizing the linear component beyond FIR filters, while maintaining computational efficiency and minimizing distortion. We also see the expansion of GBFs to accommodate multiple inputs and outputs as a crucial next step. Furthermore, we plan to investigate the potential of using the learned transformations, independent of a trained GBF, as inputs to a multi-input linear filter. This approach could facilitate iterative implementations, thereby giving practicioners more options for practical applications.

\section*{Impact Statement}

This paper presents work whose goal is to advance the field of Machine Learning. There are many potential societal consequences of our work, none which we feel must be specifically highlighted here.


\bibliography{references}
\bibliographystyle{icml2024}

\appendix
\onecolumn
\section{Additional Plots.}

\begin{figure}[ht]
\centering
\includegraphics[width=0.7\columnwidth]{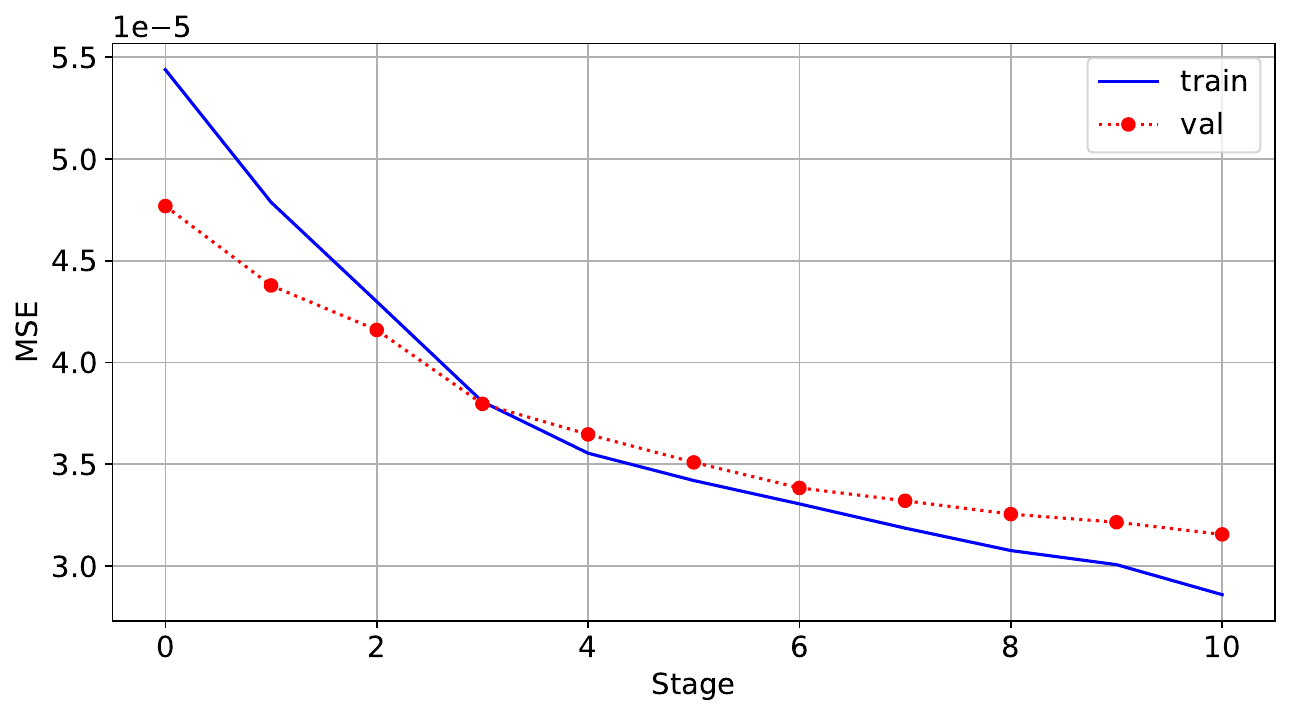}
\caption{The progressive reduction in approximation errors at each stage of the GBF model in Section \ref{sec:laptop_acoustics}.}
\label{fig:ex2_gbf_stagewise_mse}
\end{figure}

\begin{figure}[ht]
\centering
\includegraphics[width=0.7\columnwidth]{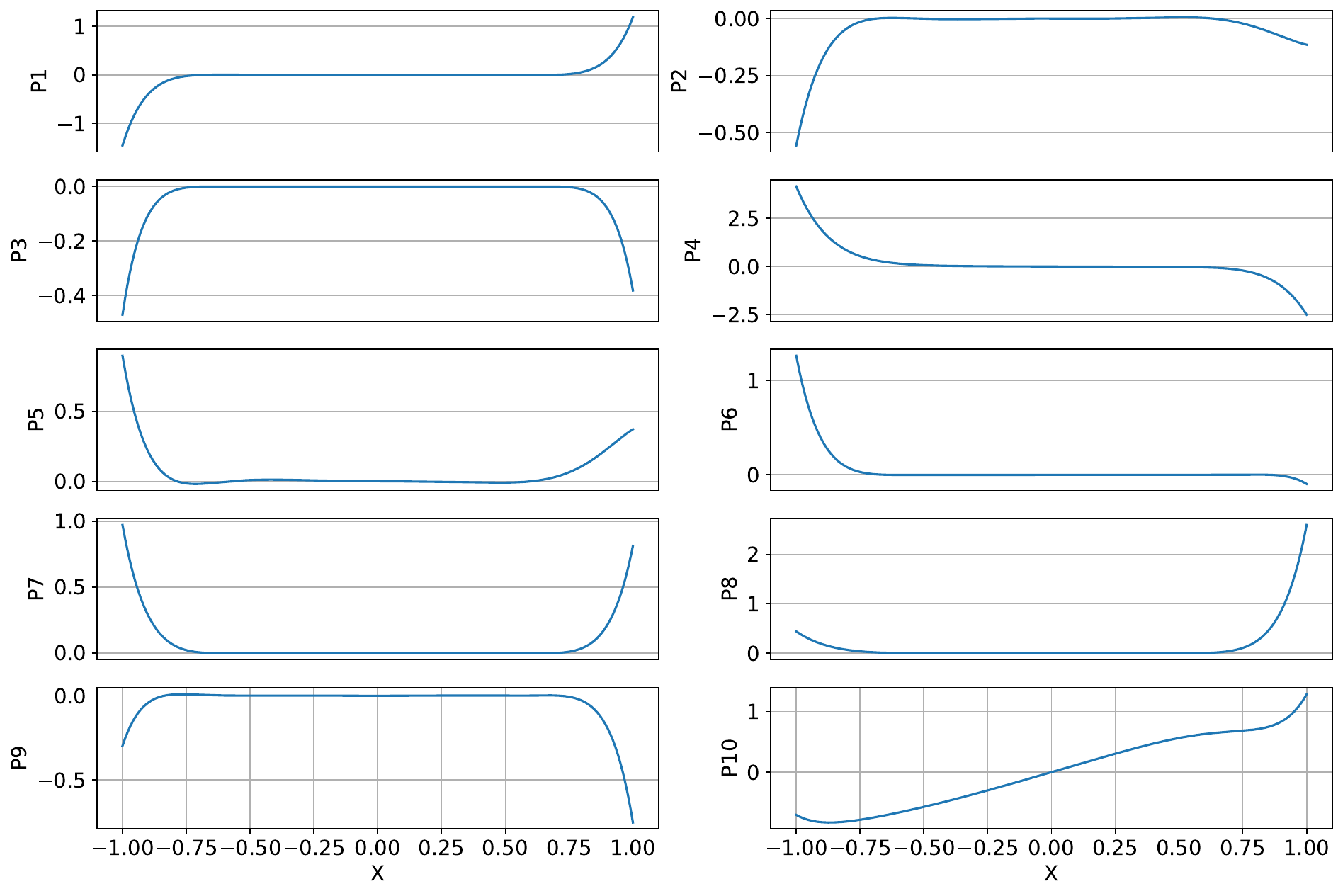}
\caption{The learned transformations for the GBF in Section \ref{sec:laptop_acoustics}.}
\label{fig:ex2_transformations}
\end{figure}

\end{document}